\documentclass{article}
\usepackage{spconf,amsmath,graphicx}
\usepackage{subfig}
\usepackage{csquotes}
\usepackage{hyperref}
\usepackage{ctable}
\usepackage{tabularx}

\newcolumntype{Y}{>{\centering \arraybackslash}X}
\newcolumntype{b}{>{\hsize=0.38\linewidth}X}
\newcolumntype{s}{>{\centering  \arraybackslash \hsize=0.01\linewidth}X}
\newcolumntype{m}{>{\centering  \arraybackslash \hsize=0.1\linewidth}X}
\title{Improving BERT Performance for Aspect-Based Sentiment Analysis}
\name{{Akbar Karimi$\qquad$ Leonardo Rossi$\qquad$ Andrea Prati }} 
\address{University of Parma\\
	Department of Engineering and Architecture\\
	Parco Area delle Scienze 181/A, Parma, Italy\\
	\{akbar.karimi, leonardo.rossi, andrea.prati\}@unipr.it
}
\begin{document}
%
\maketitle
\begin{abstract}
	Aspect-Based Sentiment Analysis (ABSA) studies the consumer opinion on the market products. It involves examining the type of sentiments as well as sentiment targets expressed in product reviews. Analyzing the language used in a review is a difficult task that requires a deep understanding of the language. In recent years, deep language models, such as BERT  \cite{devlin2019bert}, have shown great progress in this regard. In this work, we propose two simple modules called Parallel Aggregation and Hierarchical Aggregation to be utilized on top of BERT for two main ABSA tasks namely Aspect Extraction (AE) and Aspect Sentiment Classification (ASC) in order to improve the model's performance. We show that applying the proposed models eliminates the need for further training of the BERT model. The source code is available on the Web for further research and reproduction of the results\footnote{\url{https://github.com/IMPLabUniPr/BERT-for-ABSA}}.
\end{abstract}
\begin{keywords}
	Sequence Labeling, Text Classification, Aspect-Based Sentiment Analysis, BERT Fine-tuning
\end{keywords}

\section{Introduction}

In an industry setting, it is extremely important to have a valid conception of how consumers perceive the products. They communicate their perception through their comments on the products, using mostly social networks nowadays. They might have positive opinions which can lead to the success of a business or negative ones possibly leading to its demise. Due to the abundance of these views in many areas, their analysis is a time-consuming and labor-intensive task which is why a variety of machine learning techniques such as Support Vector Machines (SVM) \cite{cortes1995support, kiritchenko2014nrc}, Maximum Entropy \cite{jaynes1957information, nigam1999using}, Naive Bayes \cite{duda1973pattern, gamallo2014citius}, and Decision Trees \cite{quinlan1986induction, wakade2012text} have been proposed to perform opinion mining. 

In recent years, Deep Learning (DL) techniques have been widely utilized due to the increase in computational power and the huge amount of freely available data on the Web. One of the areas on which these techniques have had a great impact is Natural Language Processing (NLP) where modeling (i.e. understanding) the language plays a crucial role. BERT \cite{devlin2019bert} is a state-of-the-art model of this kind  which has become widely utilized in many NLP tasks \cite{kantor2019learning, davison2019commonsense} as well as in other fields \cite{peng2019transfer, alsentzer2019publicly}. It has been trained on a large corpus of Wikipedia documents and books in order to \textit{learn} the language syntax and semantics from the context. The main component of its architecture is called the transformer \cite{vaswani2017attention} block consisting of attention heads. These heads have been designed to pay particular attention to parts of the input sentences that correspond to a particular given task \cite{vig2019analyzing}. In this work, we utilize BERT for Aspect-Based Sentiment Analysis (ABSA) tasks.

Our main contribution is the proposal of two simple modules that can help improve the performance of the BERT model. In our models we opt for Conditional Random Fields (CRFs) for the sequence labeling task which yield better results. In addition, our experiments show that training BERT for more number of epochs does not cause the model to overfit. However, after a certain number of training epochs, the learning seems to stop. 

\section{Related Work}
Recently, there has a large body of work which utilizes the BERT model for various tasks in NLP in general such as text classification \cite{sun2019fine}, question answering \cite{yang2019data}, summarization \cite{liu2019fine} and, in particular, ABSA tasks. 

 Using Graph Convolutional Networks (GCNs), in \cite{zhao2020modeling}, the authors take into account sentiment dependencies in a sequence. In other words, they show that when there are multiple aspects in a sequence, the sentiment of one of them can affect that of the other one. Making use of this information can increase the performance of the model. Some studies convert the AE task into a sentence-pair classification task.  For instance, authors of \cite{sun2019utilizing} construct auxiliary sentences using the aspect terms of a sequence. Then, utilizing both sequences, they fine-tune BERT on this specific task. 

Word and sentence level representations of a model can also be enriched using domain-specific data. Authors of \cite{xu2019bert} show this by post-training the BERT model, which they call BERT-PT, on additional restaurant and laptop data. In our experiments, we use the embeddings from their work for the initialization of our models. 

Due to the particular architecture of the BERT model, extra modules can be attached on top of it. In \cite{li2019exploiting}, the authors add different layers such as an RNN and a CRF layer to perform ABSA in an end-to-end fashion. In our work, we use the same layer modules from the BERT architecture and employ the hidden layers for prediction as well. 

\section{Aspect-Based Sentiment Analysis Tasks}

Two of the main tasks in ABSA are Aspect Extraction (AE) and Aspect Sentiment Classification (ASC). We briefly describe them in this section. 

\textbf{Aspect Extraction.}
In AE, the goal is to extract a specific aspect of a product toward which some sentiment is expressed in a review. For instance, in the sentence, \textquote{\textit{The laptop has a good battery.}}, the word \textit{battery} is the aspect which is extracted. This task can be seen as a sequence labeling task, where the words are assigned a label from the set of three letters namely \{\textit{B}, \textit{I}, \textit{O}\}.  Each word in the sequence can be the beginning word of aspect terms (\textit{B}), among the aspect terms (\textit{I}), or not an aspect term (\textit{O}). 

\textbf{Aspect Sentiment Classification.}
In ASC, the goal is to extract the sentiment expressed in a review by the consumer. Given a sequence, one of the three classes of \textit{Positive}, \textit{Negative}, and \textit{Neutral} is extracted as the class of that sequence. The representation for this element is embodied in the architecture of the BERT model. For each sequence as input, there are two extra tokens that are used by the BERT model:
\begin{center}
	$[CLS], w_1, w_2, ..., w_n, [SEP]$
\end{center}
The sentiment of a sentence is represented by the $[CLS]$ token representation in the final layer of the architecture. The class probability is, then, computed by the softmax function. 

\section{Proposed Models}
\begin{figure}
	\centering
	\includegraphics[scale=0.5]{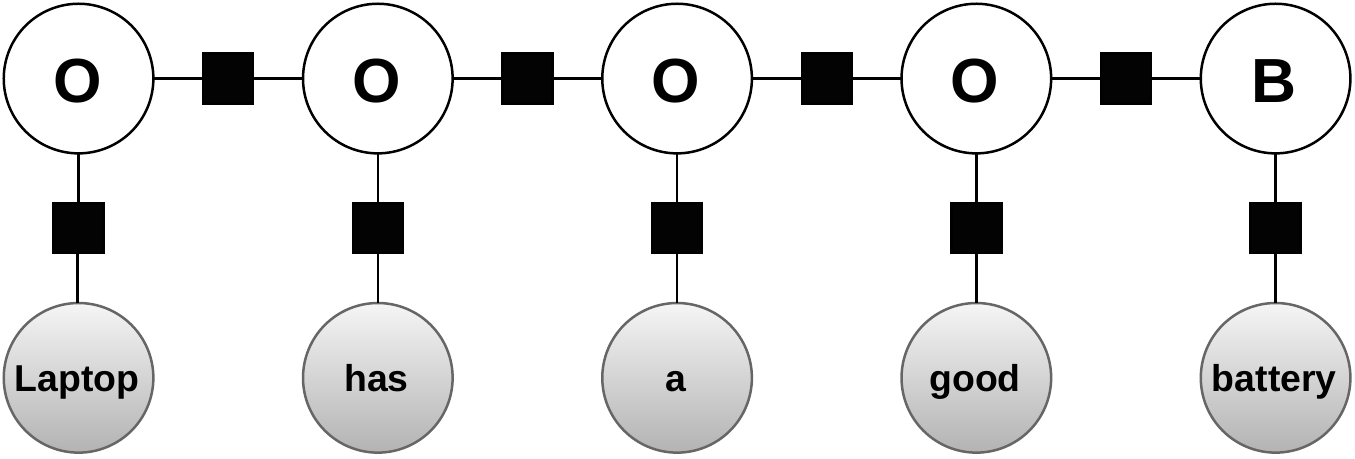}
	\caption{An example of representing a sentence with its word labels using CRFs.}
	\label{crf}
\end{figure}

Deep models can capture deeper knowledge of the language as they grow. As shown in \cite{jawahar2019does}, the initial to middle layers of BERT can extract syntactic information, whereas the language semantics are represented in higher layers. Since extracting the sentence sentiment is semantically demanding, we expect to see this in higher layers of the network. This is the intuition behind our models where we exploit the final layers of the BERT model. 

The two models that we introduce here are similar in principle, but slightly differ in implementation. Also, for the two tasks, the losses are computed differently. While for the ASC task we utilize cross-entropy loss, for the AE task, we make use of CRFs. The reason for this choice is that the AE task can be treated as sequence labeling. Therefore, taking into account the previous labels in the sequence is of high importance, which is exactly what the CRF layer does. 	

\textbf{Conditional random fields}. CRFs \cite{lafferty2001conditional} are a type of graphical models and have been used both in computer vision (e.g. for pixel-level labeling \cite{zheng2015conditional}) and in NLP for sequence labeling. Since AE can be considered a sequence labeling task, we opt for using a CRF layer in the last part of our models. The justification for the use of a CRF module for AE is that doing so helps the network to take into account the joint distribution of the labels. This can be significant since the labels of sequence words are dependent on the words that appear before them. For instance, as is seen in Figure \ref{crf}, the occurrence of the adjective \textit{good} can give the model a clue that the next word is probably not another adjective. The equation with which the joint probability of the labels is computed is as follows:
\begin{equation}
p(\textbf{y}|\textbf{x}) = \frac{1}{Z(\textbf{x})} \prod_{t=1}^{T} exp \bigg\{\sum_{k=1}^{K}\theta_k f_k(y_t, y_{t-1}, \textbf{x}_t)\bigg\}
\label{formula1}
\end{equation}

The relations between sequence words are represented by using feature functions, \{$f_k$\} in Equation \ref{formula1}. These relations can be strong or weak, or non-existent at all. They are controlled by their weights $\{\theta_k\}$ which are computed during the training phase. 

\subsection{Parallel aggregation}
\begin{figure}
	\begin{center}
		\includegraphics[scale=0.35]{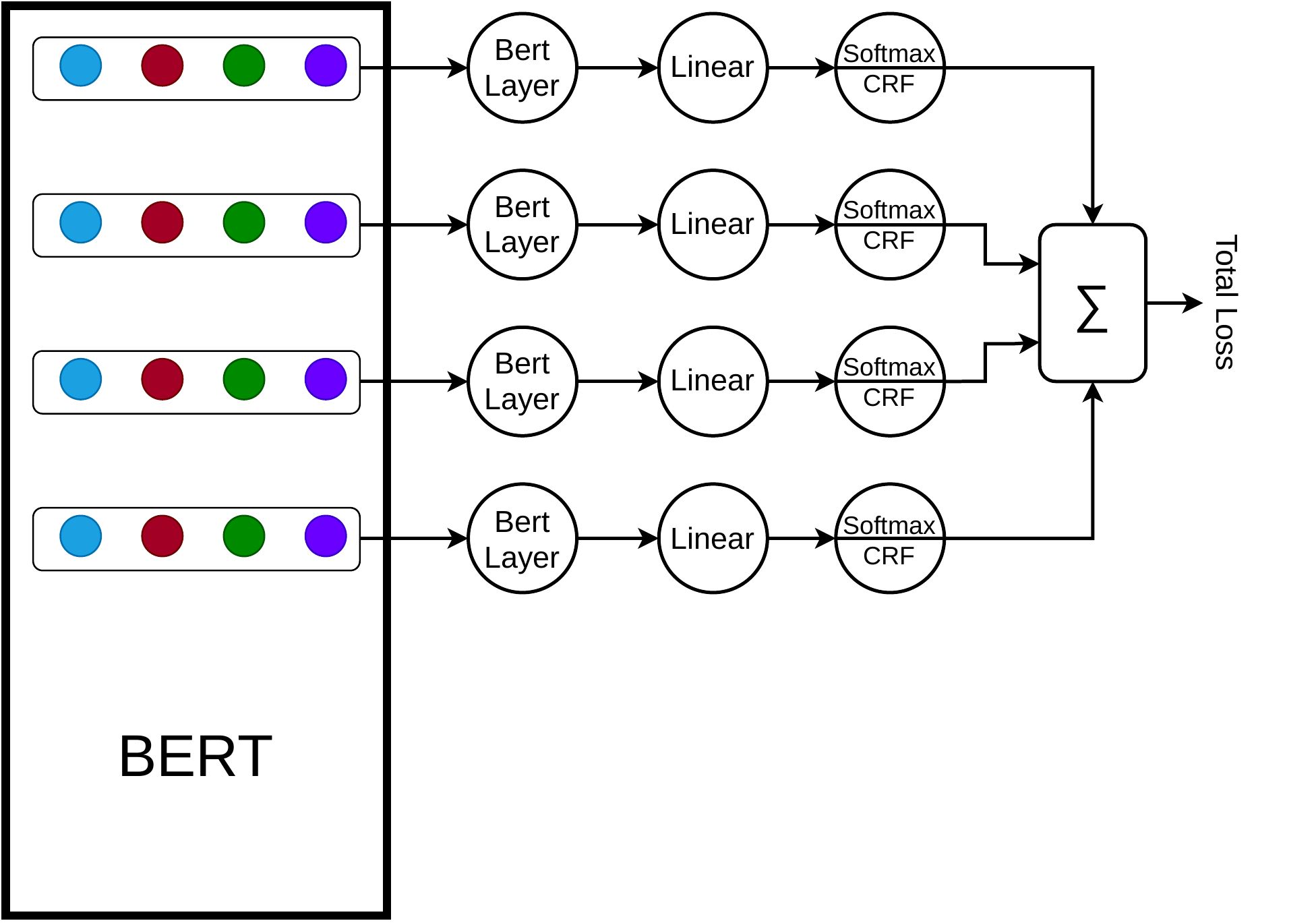}
		\caption{Parallel aggregation (P-SUM)}
		\label{modela}
	\end{center}
\end{figure}

Authors of \cite{rossi2020novel} showed that the hidden layers of deep models can be exploited more to extract region specific information. Inspired by their work, we propose parallel aggregation called P-SUM using BERT layer modules. Figure \ref{modela} shows the details of this model. We exploit the last four layers of the BERT model by adding one more BERT layer and performing prediction using each one of the layers. The reason is that all deeper layers contain most of the related information regarding the task. Therefore, extracting this information from each one of them and combining them can produce richer representations of the semantics. 

\subsection{Hierarchical aggregation}

\begin{figure}[t]
	\centering
	\includegraphics[scale=0.35]{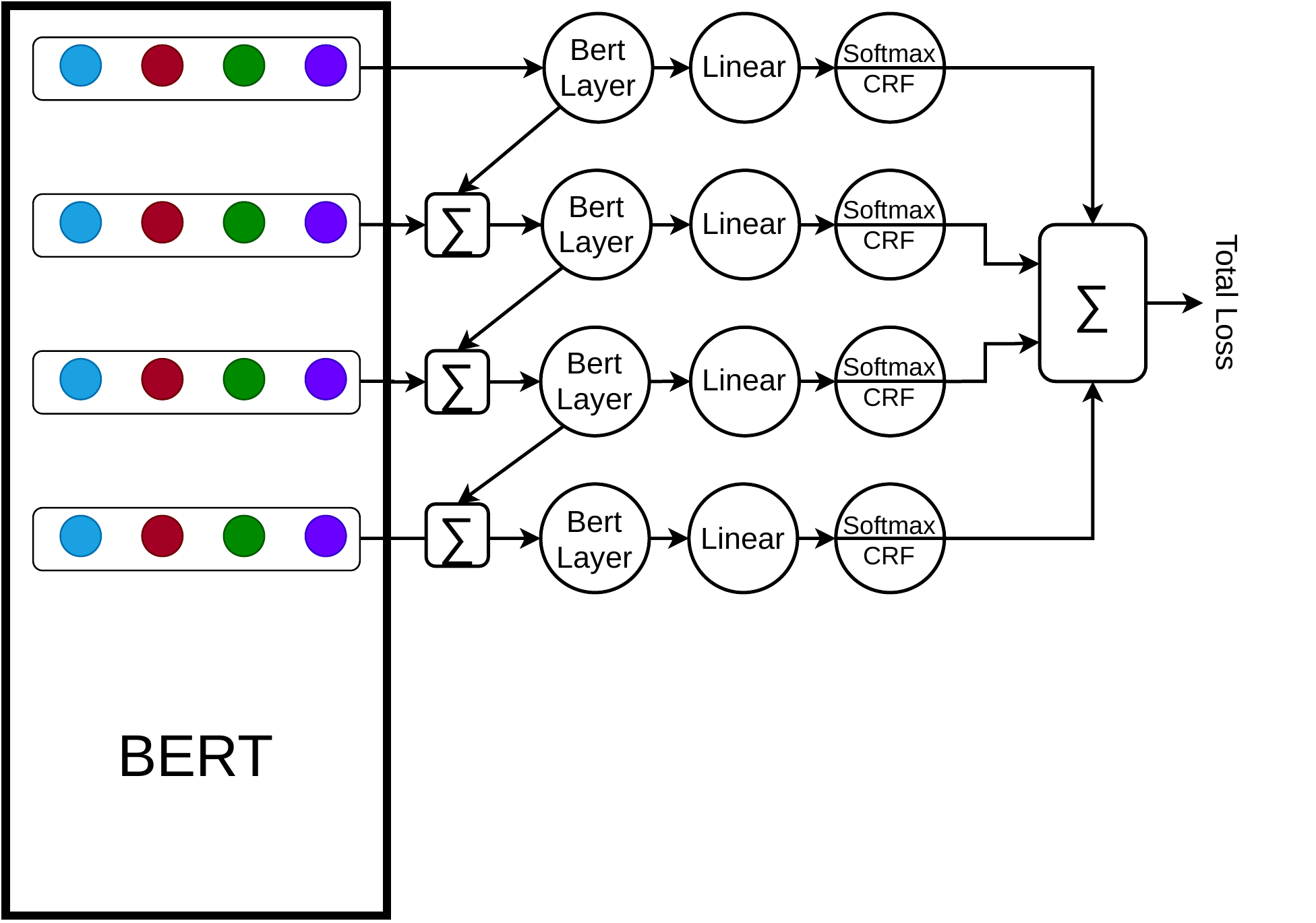}
	\caption{Hierarchical aggregation (H-SUM)}
	\label{modelb}
\end{figure}

Our hierarchical aggregation (H-SUM) model is inspired by the use of Feature Pyramid Networks (FPNs) \cite{lin2017feature}. The goal is to extract more semantics from the hidden layers of the BERT model. The architecture of the H-SUM model can be seen in Figure \ref{modelb}. Here, after applying a BERT layer on each one of the hidden layers, they are aggregated with the previous layer. At the same time, similar to the P-SUM, we perform prediction using each output branch after which the losses are also summed. 

\section{Experiments and results}

\begin{table}
	\begin{center}
		\begin{tabularx}{\columnwidth}{@{}l|Y|Y|Y|Y@{}} \hline
			& \multicolumn{2}{c|}{Train} & \multicolumn{2}{c}{Test}\\
			\hline
			\textbf{Dataset} & S & A & S & A\\
			\hline
			LPT14 & 3045 & 2358 & 800 & 654 \\
			\hline
			RST16 & 2000 & 1743 & 676 & 622\\
			\hline
		\end{tabularx}
	\end{center}
	
	\caption{Laptop (LPT14) and restaurant (RST16) datasets from SemEval 2014 and 2016, respectively, for AE. S: Number of sentences; A: Number of aspects.}
	\label{tab:ae}
	\begin{center}
		\begin{tabularx}{\columnwidth}{@{}l|Y|Y|Y|Y|Y|Y@{}} \hline
			& 
			\multicolumn{3}{c|}{Train}
			&
			\multicolumn{3}{c}{Test}\\
			\hline
			\textbf{Dataset} & Pos & Neg & Neu & Pos & Neg & Neu\\
			\hline
			LPT14 & 987 & 866 & 460 & 341 & 128 & 169 \\
			\hline 
			RST14 & 2164 & 805 & 633 & 728 & 196 & 196 \\
			\hline
		\end{tabularx}
	\end{center}
	\caption{Laptop (LPT14) and restaurant (RST14) datasets from SemEval 2014 for ASC. Pos, Neg, Neu: Number of positive, negative, and neutral sentiments, respectively.}
	\label{tab:asc}
\end{table} 

In order to carry out our experiments, we use the same codebase as \cite{xu2019bert}. We ran the experiments on a GPU (GeForce RTX 2070) with 8 GB of memory using batches of 16 for both our models and the BERT-PT model as the baseline. For training, Adam optimizer was used and the learning rate was set to $3e-5$. From the distributed training data, we used 150 examples as the validation. To evaluate the models, the official scripts were used for the AE tasks and the script from the same codebase was used for the ASC task. Results are reported in F1 for AE and in Accuracy and MF1 for ASC. 

\textbf{Datasets.} In our experiments, we utilized laptop and restaurant datasets from SemEval 2014 \cite{article} and 2016 \cite{pontiki2016semeval}. The collections consist of user reviews which have been annotated manually. The statistics of the datasets can be seen in Tables \ref{tab:ae} and \ref{tab:asc}. 

\subsection{BERT model analysis}

\begin{figure}
	\centering
	\includegraphics[scale=0.35]{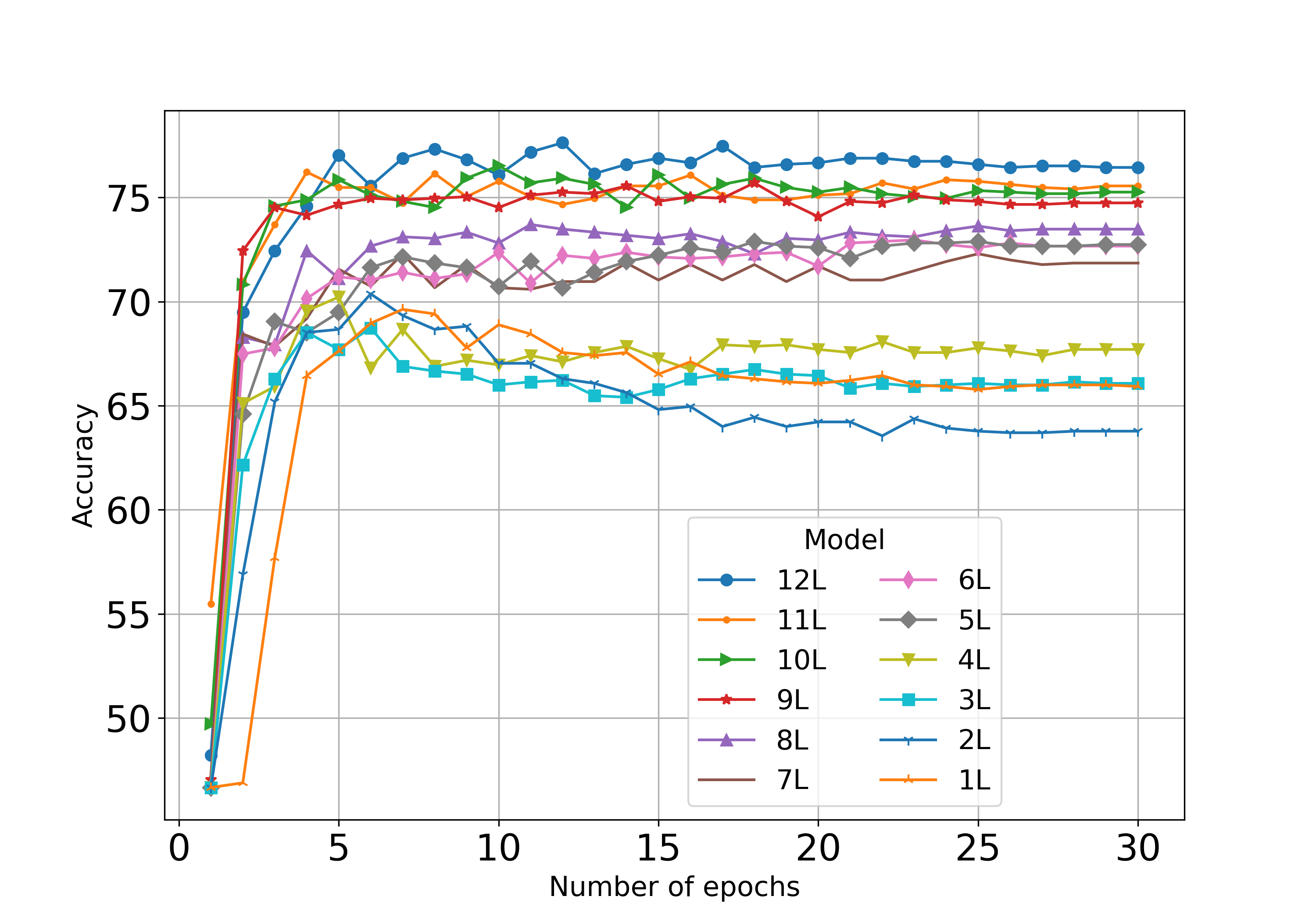}
	\caption{Performance of BERT layers for ASC on RST14 validation data.}
	\label{layers_performance}
\end{figure}

\begin{figure*}[t]
	\centering
	\subfloat[]{\includegraphics[scale=0.34]{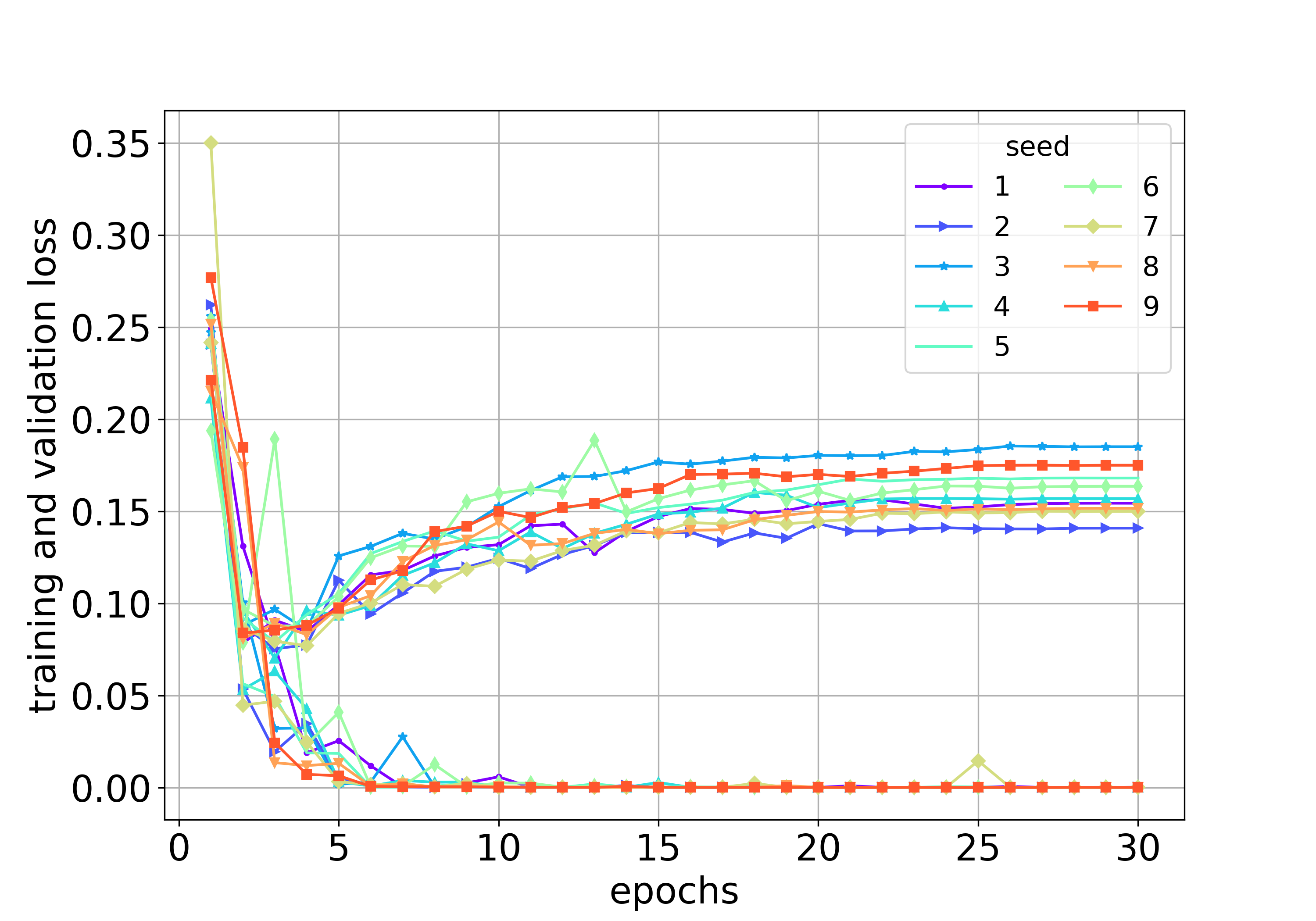}}
	\subfloat[]{\includegraphics[scale=0.34]{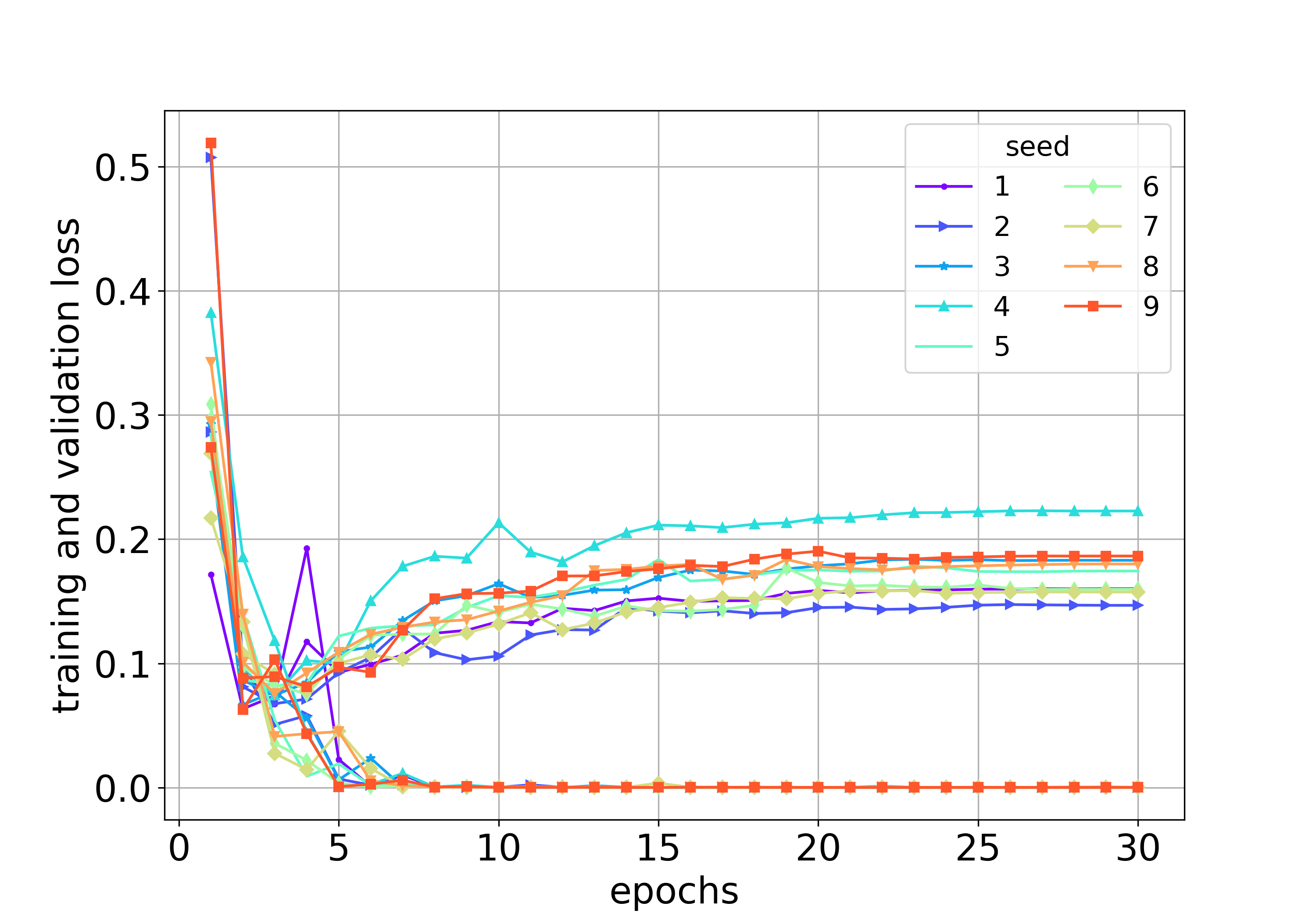}}
	
	\subfloat[]{\includegraphics[scale=0.34]{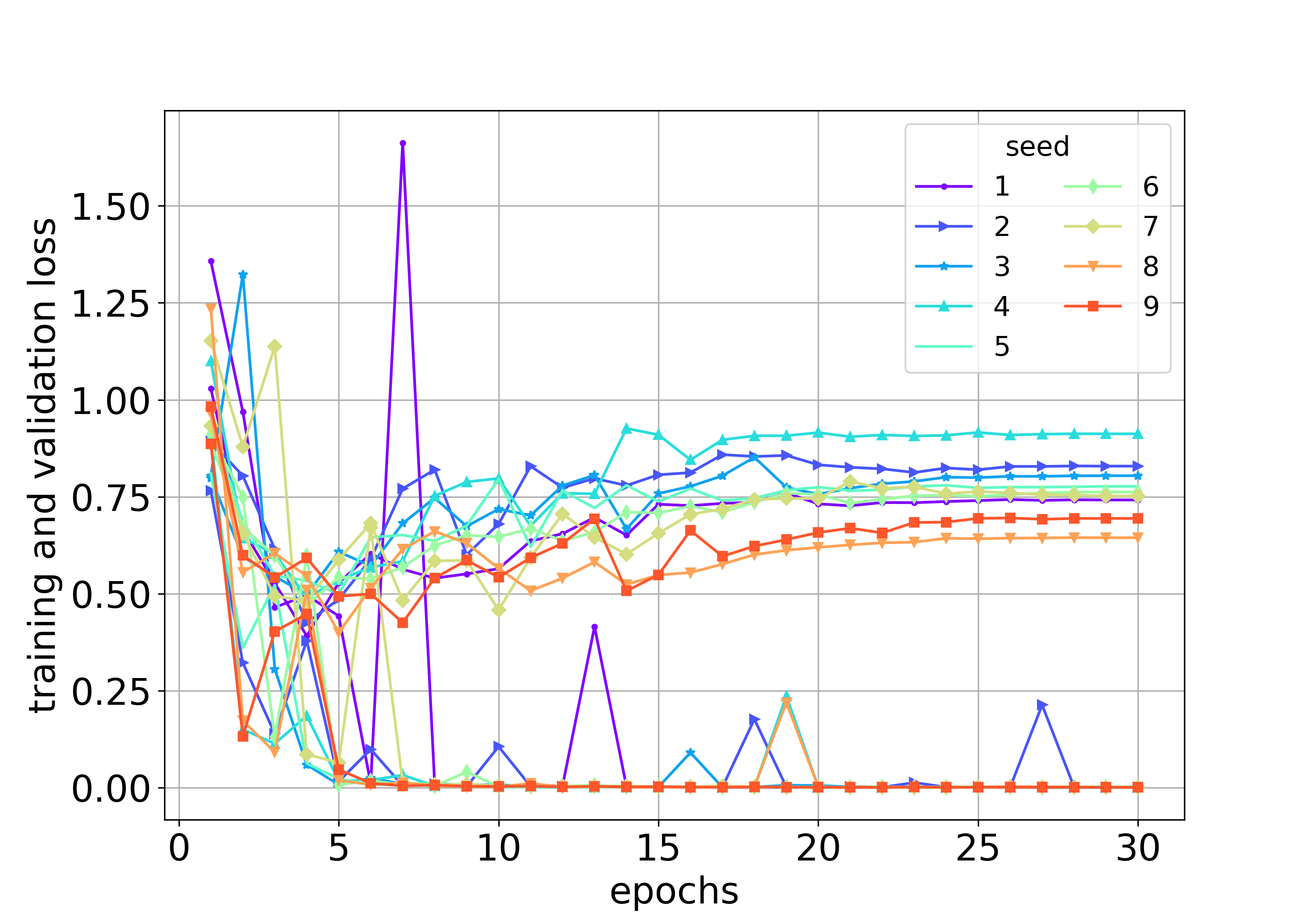}}
	\subfloat[]{\includegraphics[scale=0.34]{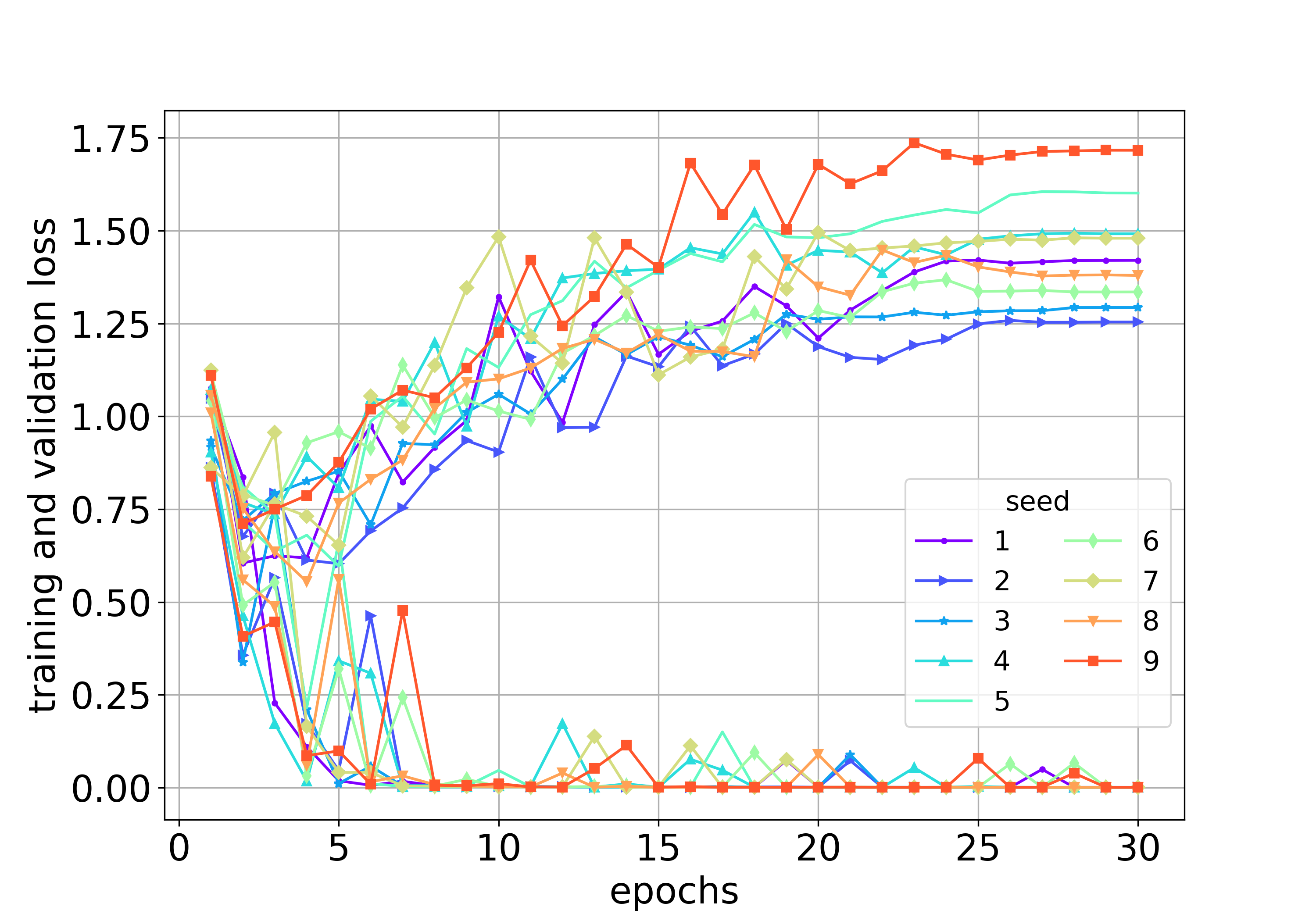}}
	\caption{Training and validation losses of BERT-PT for AE (laptop (a) and restaurant (b)) and ASC (laptop (c) and restaurant (d)). In each figure, the upper lines are validation losses and the bottom lines are training losses, each line corresponding to a seed number.}
	\label{val_losses}
\end{figure*}

\textbf{Performance of BERT layers.} 
We carried out experiments to find out how each layer of the BERT model performs. The results are shown in Figure \ref{layers_performance}. As can be seen, better performance is achieved in the deeper layers. We take the last four layers and attach our modules to them. 

\textbf{Increasing training epochs.} 
More training can lead to a better performance of the network. However, one risks the peril of overfitting especially when the number of training examples are not considered to be large compared to the number of parameters contained in the model. However, in the case of BERT, as was also observed by \cite{li2019exploiting}, it seems that with more training the model does not overfit although the number of the training data points is relatively small. The reason behind this could be the fact that we are using an already pretrained model which has seen an enormous amount of data (Wikipedia and Books Corpus). Therefore, we can expect that by performing more training, the model will still be able to generalize. 

The same observation can be made by looking at the validation losses in Figure \ref{val_losses}. In case of an overfit, we would expect the losses to go up and the performance to go down. However, we see that with the increase in loss, the performance improves as well (Figure \ref{layers_performance}).  This suggests that with more training, the network weights continue to change, which is between 15 to 20 epochs,  after which they remain almost stable indicating that there is no more learning.

\subsection{Results}

\begin{table*}
	\caption{Comparison of results. \textit{BERT-PT*} is the original BERT-PT model using our model selection. The boldfaced numbers show the outperforming models using the same settings. The underlined numbers indicate where more training can be better. Each score in the table is the average of 9 runs. Acc: Accuracy, MF1: Macro-F1.}
	\begin{center}
		\begin{tabularx}{0.8\linewidth}{b m m s m m s m m}
			\specialrule{.15em}{.1em}{.1em}
			& \multicolumn{2}{c}{AE} & & \multicolumn{5}{c}{ASC} \\
			\cline{2-3} \cline{5-9}
			& Laptop & Rest16 & & \multicolumn{2}{c}{Laptop} & &\multicolumn{2}{c}{Rest14} \\
			\cline{5-6} \cline{8-9}
			\textbf{Models}  & F1 & F1 & & Acc & MF1 & & Acc & MF1 \\
			\hline
			BERT & 79.28 & 74.10 & & 75.29 & 71.91 & & 81.54 & 71.94\\
			\hline
			DE-CNN \cite{xu2018double} & 81.59 & 74.37 & & - & - &&- & -\\
			BERT-PT \cite{xu2019bert} & 84.26 & 77.97  & & 78.07 & 75.08 & & 84.95 & 76.96 \\
			BAT \cite{karimi2020adversarial} & 85.57 & 81.50 & & 79.35 & 76.50 & & 86.03 & 79.24 \\
			BERT-PT (30 epochs) & 85.93 & \textit{\underline{82.64}} && \textit{\underline{79.48}} & 76.47&&86.09 & 79.06\\
			BERT-PT* (4 epochs) & 85.57 & 81.57 & & 78.21 & 75.03 & & 85.43 & 77.68\\
			P-SUM (4 epochs)& \textbf{85.94} &  \textbf{81.99} & &  \textbf{79.55} &  \textbf{76.81} & &  \textbf{86.30} &  \textbf{79.68}\\
			H-SUM (4 epochs)& \textbf{86.09} & \textbf{82.34} & & \textbf{79.40} & \textbf{76.52} & & \textbf{86.37} & \textbf{79.67}\\
			\specialrule{.1em}{.1em}{.1em} 
		\end{tabularx}
	\end{center}
	\label{all}
\end{table*}

As expected, our experimental results show that with the increase of the training epochs the BERT model also improves. These results can be seen in Table \ref{all}. To compare our proposed models with \cite{xu2019bert}, we perform the same model selection for both of them. Unlike \cite{xu2019bert} and \cite{karimi2020adversarial} who select their best models based on the lowest validation loss, we choose the models trained with four epochs after observing that accuracy goes up on the validation sets (Figure \ref{layers_performance}). Therefore, in Table \ref{all}, we report the original BERT-PT scores as well as the ones for our model selection. Compared to training BERT-PT with 30 epochs, in all cases except for AE (restaurant), our models produce better results in terms of F1 and Macro-F1. From the table, it can also be seen that the proposed models outperform the newly selected BERT-PT model in both datasets and tasks with improvements in MF1 as high as \textbf{+1.78} and \textbf{+2} for ASC on latpop and restaurant, respectively. 

\section{Conclusion}

We proposed two simple modules utilizing the hidden layers of the BERT language model to produce deeper semantic representations of input sequences. The layers are once aggregated in a parallel fashion and once hierarchically. We perform prediction on each one of the selected hidden layers and compute the loss. These losses are then aggregated to produce the final loss of the model. We address aspect extraction using conditional random fields which helps take into account the joint distribution of the sequence labels to achieve more accurate predictions. Our experiments show that the proposed approaches outperform the post-trained vanilla BERT model. 

\bibliographystyle{IEEEbib}
\bibliography{strings,refs}

\end{document}